\documentclass[runningheads]{llncs}
\usepackage{xcolor}

\usepackage[T1]{fontenc}
\usepackage{graphicx}
\usepackage{verbatim}
\usepackage{multirow}
\usepackage{makecell}
\usepackage{amsmath, amssymb}
\usepackage{hyperref}
\begin{document}
\title{Memory-Augmented SAM2 for Training-Free Surgical Video Segmentation}
\titlerunning{MA-SAM2}

\author{Ming Yin* \and Fu Wang\and Xujiong Ye \and Yanda Meng \and Zeyu Fu*} 

% index{Yin, Ming}
% index{Wang, Fu}
% index{Ye, Xujiong}
% index{Meng, Yanda}
% index{Fu, Zeyu}

%% Added for anonymized MICCAI 2025 submission
\authorrunning{M.Yin et al.}
\institute{ Department of Computer Science, University of Exeter, UK \\
    \email{\{my417, Z.Fu\}@exeter.ac.uk}}
\maketitle              % typeset the header of the contribution
\begin{abstract}
Surgical video segmentation is a critical task in computer-assisted surgery, essential for enhancing surgical quality and patient outcomes. Recently, the Segment Anything Model 2 (SAM2) framework has demonstrated remarkable advancements in both image and video segmentation. However, the inherent limitations of SAM2's greedy selection memory design are amplified by the unique properties of surgical videos—rapid instrument movement, frequent occlusion, and complex instrument-tissue interaction—resulting in diminished performance in the segmentation of complex, long videos. To address these challenges, we introduce Memory Augmented (MA)-SAM2, a training-free video object segmentation strategy, featuring novel context-aware and occlusion-resilient memory models. MA-SAM2 exhibits strong robustness against occlusions and interactions arising from complex instrument movements while maintaining accuracy in segmenting objects throughout videos. Employing a multi-target, single-loop, one-prompt inference further enhances the efficiency of the tracking process in multi-instrument videos. Without introducing any additional parameters or requiring further training, MA-SAM2 achieved performance improvements of 4.36\% and 6.1\% over SAM2 on the EndoVis2017 and EndoVis2018 datasets, respectively, demonstrating its potential for practical surgical applications. The code is available at \url{https://github.com/Fawke108/MA-SAM2}.

\end{abstract}
\section{Introduction}

Surgical instrument segmentation (SIS), a critical component of medical imaging and surgical scene analysis, aims to achieve pixel-level localization and contour extraction of instruments from intraoperative videos~\cite{jian2020multitask,yue2023cascade,zhang2020empowering,shademan2016supervised}. Despite the advancements in deep learning-based SIS methods, which have demonstrated high segmentation accuracy and robustness, these methods often depend on extensive expert annotations and encounter significant challenges in achieving efficient real-time segmentation of surgical videos~\cite{yang2023tma,ayobi2023matis,song2025attention,ni2022surginet,yang2022attention}.

The Segment Anything Model (SAM)~\cite{kirillov2023segment}, a foundational model for promptable image segmentation, has revolutionised the field by enabling robust zero-shot generalisation across diverse segmentation tasks. Leveraging user-provided prompts, SAM generates precise segmentation masks for objects within static images. This remarkable zero-shot learning capability has been validated by numerous studies~\cite{williams2024leaf,peng2025sam}, as exemplified by PerSAM~\cite{zhang2023personalize}, which efficiently adapts SAM to specific objects or styles using only a single reference image. Recent research has further explored SAM's potential in the medical domain, including in surgical instrument segmentation~\cite{ma2024segment,wu2023medical,yue2024surgicalsam}. However, SAM’s initial design is tailored for static images, limiting its effectiveness in video segmentation tasks where temporal consistency is essential. 

To address this limitation, SAM2~\cite{ravi2024sam} was introduced as an extension of SAM for video segmentation tasks. By incorporating a memory mechanism that stores information from recently processed frames, SAM2 leverages temporal context to enhance segmentation consistency and accuracy across sequential video frames. Recent studies have increasingly focused on SAM2’s application in medical image and video segmentation, particularly in surgical instrument segmentation~\cite{shen2024performance,yan2024biomedical,zhu2024medical}. For instance, Surgical SAM2~\cite{liu2024surgical} introduced an efficient frame pruning strategy which analyses inter-frame similarity and information content to selectively process key frames, reducing redundant computations. While these approaches have shown some effectiveness, they are not training-free and fail to fully resolve the reasoning errors in SAM2 caused by its greedy memory selection design.

\begin{figure}[t]
\centering
\includegraphics[width=0.8\textwidth]{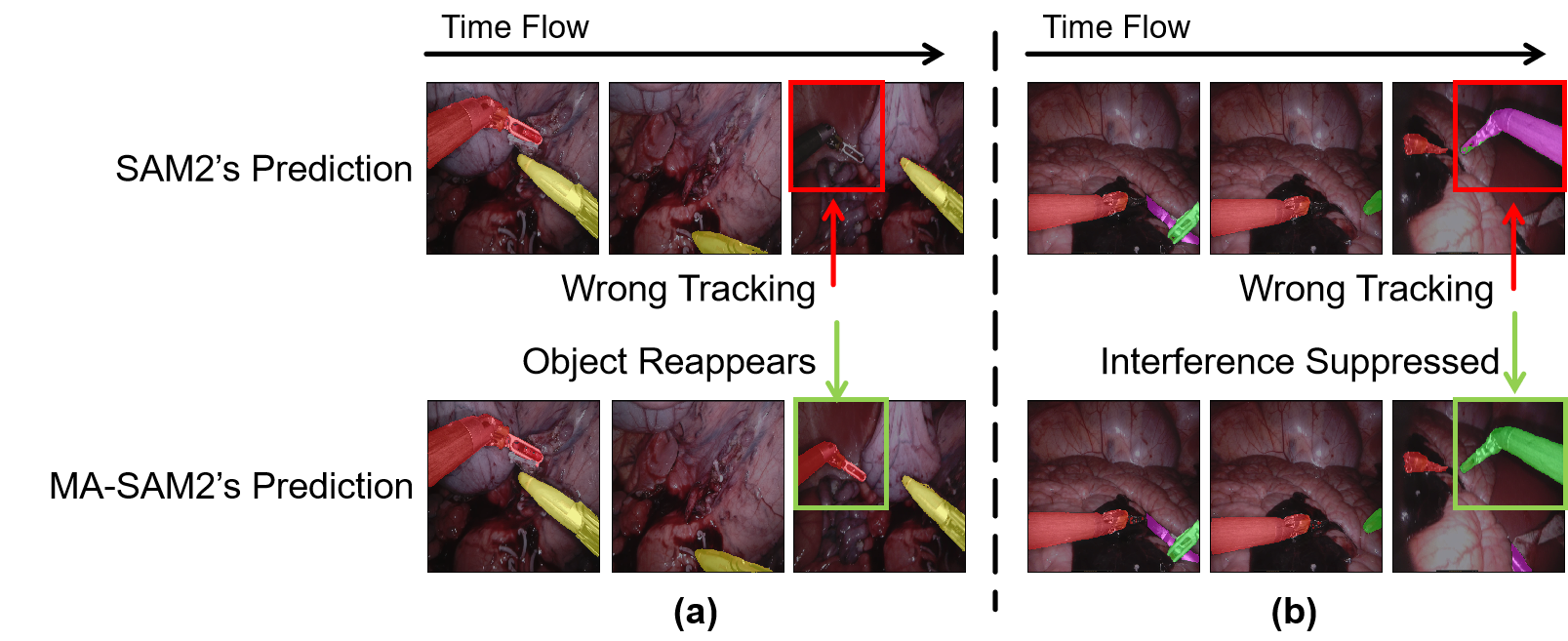}
\caption{Performance comparison of occlusion handling and temporal consistency between MA-SAM2 and SAM2: (a) Tracking over time during instrument reappearance, (b) Tracking over time during multiple instrument overlap.} \label{fig1}
\end{figure}

Our empirical findings show that SAM2's memory bank, which processes frames sequentially based on temporal order, accumulates instability when uncertain or low-confidence masks are generated in intermediate frames. As depicted in Fig.~\ref{fig1}(a), SAM2 exhibits tracking failure when an instrument disappears and subsequently reappears. Similarly, Fig.~\ref{fig1}(b) illustrates the occurrence of segmentation ambiguities in SAM2 during instances of multiple instrument overlap.

\begin{figure}[t]
\centering
\includegraphics[scale=0.3]{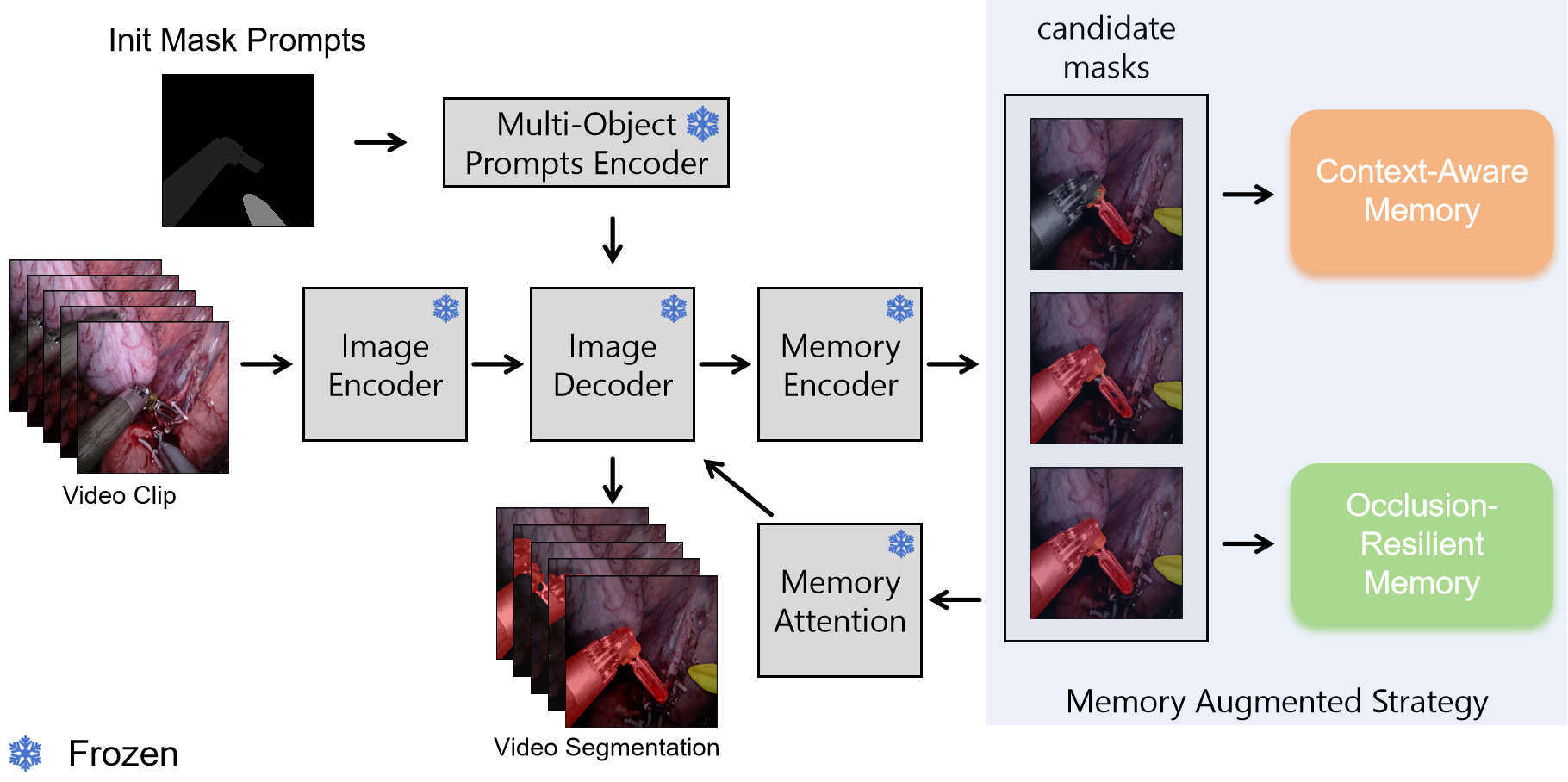}
\caption{Architecture of the proposed model MA-SAM2.} \label{fig2}
\end{figure}

To address the aforementioned challenges, we propose Memory Augmented (MA)-SAM2, a training-free video segmentation strategy, as shown in Fig.~\ref{fig2}. We design a memory augmented strategy to dynamically integrate context-aware memory (CAM) and occlusion-resilient memory (ORM). Specifically, CAM utilises a collaborative hypothesis pruning mechanism to maintain optimal target clues during long-term video inference, thereby mitigating the impact of erroneous predictions and resolving the issue of lost reappearance identification. ORM employs a variation selection mechanism to identify and suppress intra-frame interference information, thereby handling the scenario of multiple overlapping instruments causing feature contamination. Simultaneously, considering the real-time requirements of surgical video segmentation, we adopt a one-prompt strategy. Unlike methods that process one target at a time, MA-SAM2 performs simultaneous multi-target inference using a mask-based one-prompt strategy provides a single prompt upon each category’s first appearance, no further manual intervention or corrective prompts are needed throughout the sequence. This design facilitates efficient inference across long and complex surgical videos.

In summary, our contributions are as follows:
1) We propose MA-SAM2, a training-free surgical instrument tracking approach that improves upon SAM2 through a one-prompt mechanism. This approach mitigates segmentation jitters caused by instrument displacements and segmentation ambiguities resulting from instrument interactions, enhancing efficiency in surgical videos. 2) We propose a new memory bank architecture, which dynamically incorporates CAM and ORM, enhancing long-term temporal modelling, ensuring robust tracking despite occlusion interference. 3) Experiments on EndoVis2017 and EndoVis2018 show that MA-SAM2 outperformed SAM2 by 6.1\% and 4.36\% in Challenge IoU, respectively, demonstrating its superior performance in surgical instrument segmentation.

\section{Method}
\subsection{Problem Formulation}

Given a surgical video sampled into \( N \) consecutive frames, where each frame at time step \( t \in \{1, 2, ..., N\} \) is denoted as \( I_t \in \mathbb{R}^{H \times W \times 3} \), with $H \times W$ representing the spatial dimensions of the image embedding. The goal is to generate a segmentation mask \( M(c, t) \) for each instrument category \( c \in C \) at frame \( t \). The segmentation is performed by a model \( e.g., \text{SAM2} \), formulated as:
\begin{equation}
M(c, t) = \text{SAM2}(\{I_1, ..., I_N\}, c, t)
\end{equation}

As shown in Fig \ref{fig2}, the SAM2 process each input frame through a hierarchical image encoder to extract multi-scale features from streaming video frames. The prompts are encoded and combined with these features to guide segmentation. A memory encoder fuses past segmentation masks with frame embeddings and stores both prompted frames and a FIFO queue of recent unprompted instances in a memory bank. Memory attention layers then condition current features via self-attention and cross-attention, enhancing temporal consistency. Finally, a mask decoder assesses object presence and also predicts precise masks.

However, the FIFO mechanism used in SAM2’s memory bank leads to sequential memory, which can accumulate instability due to uncertain masks. To address this, MA-SAM2 adopts a memory-augmented strategy that dynamically integrates context-aware and occlusion-resilient memories. This results in more stable and comprehensive segmentation representations, effectively overcoming the memory limitations observed in SAM2.

\begin{figure}[t]
\centering
\includegraphics[width=\textwidth]{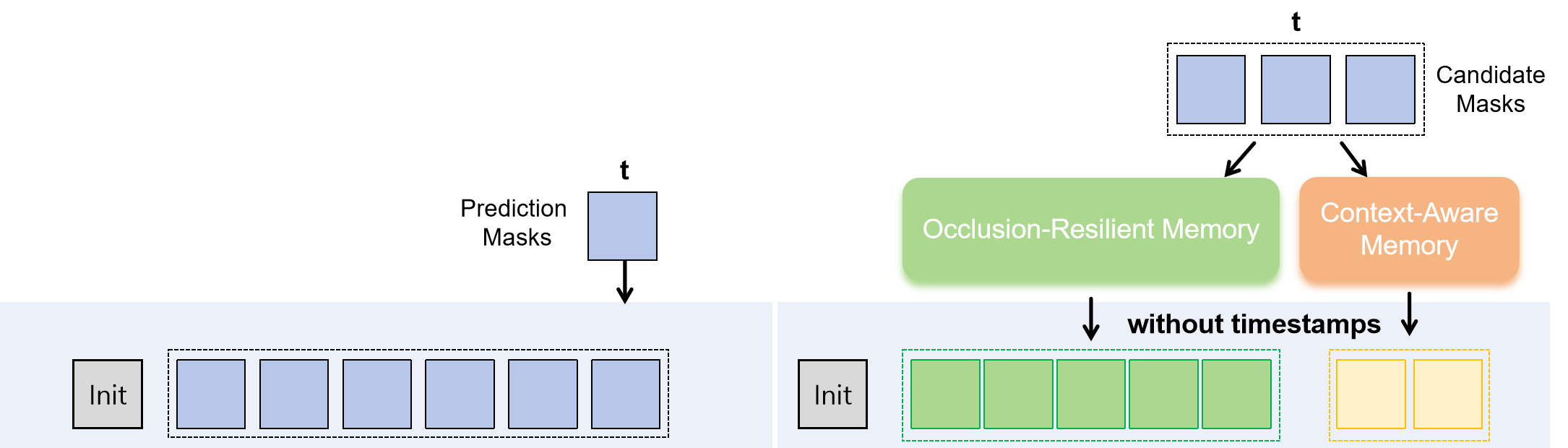}
\caption{Memory bank comparison between the SAM2 and our MA-SAM2.} \label{fig3}
\end{figure} 

\subsection{MA-SAM2} 
\subsubsection{Memory Augmented Strategy}

The memory-augmented strategy separates candidate masks into CAM and ORM, as illustrated in Fig.~\ref{fig3}, maintaining distinct memory portions within a fixed-capacity bank. Candidate masks from the image decoder are simultaneously fed into both branches: CAM employs collaborative hypothesis pruning, comparing accumulated branch values from paired masks of current and previous frames to preserve the most plausible segmentation, while ORM evaluates IoU and confidence scores, applying a variation selection mechanism to capture occlusion-related information and store the optimal mask.

To accommodate dynamic video content, the bank ensures ORM is populated first, with CAM added only if additional capacity remains. This adaptive design dynamically balances CAM and ORM, integrating their outputs to build a more stable and comprehensive segmentation representation, thereby enhancing performance in the complex context of surgical videos.

\subsubsection{Occlusion-Resilient Memory} 
The occlusion-resilient memory is designed to suppress segmentation artefacts arising from instrument motion, disappearance, or occlusion. By analysing inter-frame dynamics, it rectifies temporal instabilities, thereby enhancing the temporal consistency and robustness of the segmentation.

Inspired by Videnovic et al.~\cite{videnovic2024distractor}, we extend its application to multi-target simultaneous inference. Specifically, we generate candidate masks $M_i$ containing multiple targets per frame, each representing a different segmentation hypothesis in a given frame and potentially containing multiple instrument instances. These masks are evaluated using the average IoU score for targets present in the current frame. Among them, the mask with the highest IoU score is selected as the primary prediction, denoted as $M_s$. The remaining candidate masks are considered alternative hypotheses, denoted collectively as $M_a$. To ensure the validity of $M_s$, we impose a threshold $\theta_{IoU} \geqslant 0.8$ to exclude uncertain predictions. Then, to refine the interference regions and suppress noise, we post-process the alternative masks by removing regions that overlap with $M_s$ and retaining only their largest connected components. This produces a final refined interference mask $M_f$, computed as:
\begin{equation}
M_f=CC(M_s \setminus (M_s\cap M_a))\cup (M_s \cap M_a)
\end{equation}
where $M_s \cap M_a$ denotes the overlapping region between the primary mask and the union of alternative masks, $M_s \setminus (M_s\cap M_a)$ extracts the non-overlapping part of $M_s$, $CC(\cdot)$ denotes extracting the largest connected component of the mask. We then assess the spatial consistency between $M_f$ and $M_s$. If their bounding box overlap ratio falls inside a pre-defined range of 0.6 to 0.9, we conclude that there is no significant interference in the frame, and thus accept and store the current result. This strategy prevents the memory from being updated with ambiguous masks that might undermine tracking stability, and simultaneously strengthens its responsiveness to variations in objects.

We extract frames with strong interference information identified during the filtering process and store them in the occlusion-resilient memory. To ensure efficient inference, we limit its maximum storage to 5 frames, with stored content dynamically replaced during the inference process. Limiting the stored frames not only maintains inference speed by reducing memory redundancy but also prevents distant historical occlusion frames from introducing irrelevant information that could negatively impact inference.

\subsubsection{Context-Aware Memory} 
The context-aware memory (CAM) addresses the limitations imposed by SAM2's greedy short-term optimal segmentations in long surgical videos. To mitigate the cumulative effect of erroneous inferences, CAM integrates cross-temporal feature information by retaining high-quality segmentation masks from historical frames. Specifically, it filters frames not already stored in the occlusion-resilient memory and excludes the initial prompt frame to avoid redundancy and interference. For each frame within the filtering range, CAM evaluates the average confidence and IoU scores of all targets, and only when both exceed preset thresholds is the frame's segmentation mask deemed reliable and stored, up to its dynamic maximum capacity. This preserves high-quality segmentation data, mitigates accumulated errors, resolves target reappearance ambiguities, and enhances long-term tracking stability and accuracy.

Specifically, for each frame, the decoder head produces three candidate masks and computes the multi-target average IoU scores, denoted as $IoU_1(t)$, $IoU_2(t)$, $IoU_3(t)$, where $t$ is the current frame index. The system maintains a cumulative score for each branch by adding the logarithm of its current IoU to the previous total. The branch with the highest accumulated score is then selected as the optimal segmentation hypothesis over time. Inspired by Ding et al. ~\cite{ding2024sam2long}, the cumulative score for candidate mask $k$ at frame is calculated as:
\begin{equation}
S_k(t)=S(t-1)+\log{(IoU_k(t)+\epsilon)}
\end{equation}
where $k$ represents the index of the candidate mask, $t$ represents the current frame, and $\epsilon$ is a small constant to avoid taking the logarithm of zero. This process allows the system to evaluate each potential tracking hypothesis by assessing its corresponding score. We select the candidate mask with the highest cumulative score as the final result and carry it over to the next time step.

\section{Experiments}
\subsubsection{Datasets and Evaluation Metrics}
We use two datasets, EndoVis2017~\cite{Endovis2017} and EndoVis2018~\cite{Endovis2018}, for evaluation. The EndoVis2017 dataset, comprising eight video sequences of 225 frames each, was pre-processed following the approach described by Shvets et al.~\cite{shvets2018automatic}. The EndoVis2018 dataset, with 15 videos of 149 frames each. We use the instrument-type segmentation annotation provided by Cristina González et al.~\cite{gonzalez2020isinet}. 
Both datasets provide seven instrument categories. The evaluated instrument categories include Bipolar Forceps (BF), Prograsp Forceps (PF), Large Needle Driver (LND), Suction Instrument (SI), Vessel Sealer (VS), Clip Applier (CA), Grasping Retractor (GR), Monopolar Curved Scissors (MCS), and Ultrasound Probe (UP). As our approach is training-free, all videos were used for testing at the instrument-type label level~\cite{yu2024sam}.
For evaluation, we adhered to prior research and adopted three segmentation metrics: Challenge IoU~\cite{Endovis2017}, IoU, and mean class IoU (mcIoU).

\subsubsection{Implementation Details}
All experiments were conducted in a training-free setting, where the segmentation framework relied solely on the initial objects' masks in each video sequence as an initialisation prompt. This approach ensures that the evaluation reflects the model's generalisation capabilities without the need for fine-tuning or retraining. Input images were resized to $512 \times 512$.

\subsubsection{Results and Discussion}
We compared our method with TrackAnything, PerSAM, SurgicalSAM2 and SAM2 under zero-shot evaluation. As shown in Table~\ref{tab1} and Table~\ref{tab2}, our method demonstrates significant advantages in segmentation performance under zero-shot evaluation. Notably, SAM2-based models excel over TrackAnything and PerSAM, confirming SAM2's strong foundation for zero-shot medical segmentation and its memory module's efficacy in long videos.

\begin{table}[t]
\centering
\caption{Performance comparison on EndoVis2017 under zero-shot evaluation.}\label{tab1}
\resizebox{\textwidth}{!}{
\begin{tabular}{c c c c c c c c c c c }
\hline
\multirow{2}{*}{Method} & \multirow{2}{*}{Challenge IoU} & \multirow{2}{*}{IoU} & \multirow{2}{*}{mcIoU} &
\multicolumn{7}{c}{Instrument Categories} \\
\cline{5-11}
& & & & BF & PF & LND & VS & GR & MSC & UP\\
\hline
TrackAnything (1 Point)~\cite{yang2023track} & 54.90 & 52.46 & 55.35 & 47.59 & 28.71 & 43.27 & \textbf{82.75} & \textbf{63.10} & 66.46 & 55.54 \\
PerSAM~\cite{zhang2023personalize} & 42.47 & 42.47 & 41.80 & 53.99 & 25.89 & 50.17 & 52.87 & 24.24 & 47.33 & 38.16 \\
Surgical SAM2~\cite{liu2024surgical} & 54.55 & 54.55 & 54.55 & 42.21 & 46.58 & 59.18 & 50.07 & 37.39 & 88.60 & 57.04\\
SAM2~\cite{ravi2024sam} & 56.39 & 56.39 & 55.38 & 40.70 & \textbf{50.92} & \textbf{67.77} & 48.38 & 33.40 & 70.66 & 67.14\\
\hline
Ours & \textbf{62.49} & \textbf{62.49} & \textbf{59.89} & \textbf{54.41} & 50.41 & 64.73 & 73.72 & 32.66 & \textbf{72.64} & \textbf{70.85}\\
\hline
\end{tabular}}
\end{table}

Specifically, on EndoVis2017, our method achieved a Challenge IoU of 62.49\% and an mcIoU of 59.89\%, representing improvements of 6.10\% and 4.51\% over SAM2, respectively. In addition, performance in the BF, MSC, and UP categories was notably superior, indicating enhanced generalization and robustness across various instrument types.

\begin{table}[t]
\centering
\caption{Performance comparison on EndoVis2018 under zero-shot evaluation.}\label{tab2}
\resizebox{\textwidth}{!}{
\begin{tabular}{c c c c c c c c c c c }
\hline
\multirow{2}{*}{Method} & \multirow{2}{*}{Challenge IoU} & \multirow{2}{*}{IoU} & \multirow{2}{*}{mcIoU} &
\multicolumn{7}{c}{Instrument Categories} \\
\cline{5-11}
& & & & BF & PF & LND & SI & CA & MSC & UP\\
\hline
TrackAnything (1 Point) & 40.36 & 38.38 & 20.62 & 30.20 & 12.87 & 24.46 & 9.17 & 0.19 & 55.03 & 12.41 \\
PerSAM & 49.21 & 49.21 & 34.55 & 51.26 & 34.40 & 46.75 & 16.45 & 15.07 & 52.28 & 25.62 \\
Surgical SAM2 & 56.53 & 56.53 & 55.39 & 46.71 & 37.46 & \textbf{73.14} & \textbf{37.87} & 30.67 & 72.10 & 55.04 \\
SAM2 & 60.04 & 60.04 & 58.40 & 54.44 & 35.11 & 70.91 & 27.58 & 35.24 & 70.59 & 74.32\\
\hline
Ours & \textbf{64.40} & \textbf{64.40} & \textbf{62.13} & \textbf{56.93} & \textbf{44.91} & 66.73 & 36.82 & \textbf{37.65} & \textbf{75.35} & \textbf{74.51} \\
\hline
\end{tabular}}
\end{table}

On EndoVis2018, our method led in Challenge IoU 64.40\% and mcIoU 62.13\%, outperforming SAM2 by 4.36\% and 3.73\%, respectively. It showed exceptional performance in complex categories like MSC 75.35\% and PF 44.91\%. These results demonstrate that our method has superior performance in handling challenging scenes and complex instrument categories.

\begin{figure}[t]
\centering 
\includegraphics[width=0.75\textwidth]{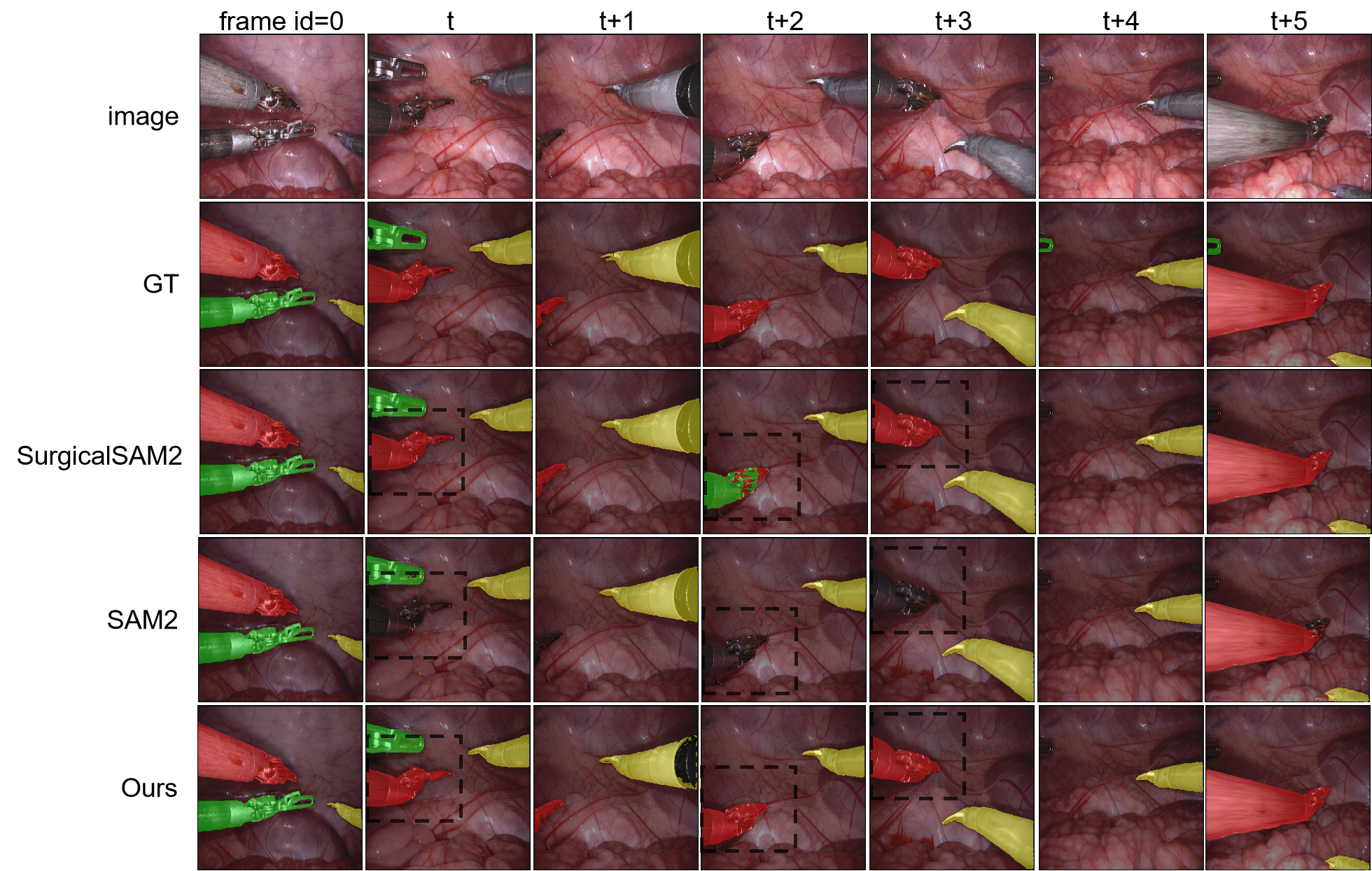}
\caption{Qualitative Comparison on the EndoVis2018 Dataset.} \label{fig6}
\end{figure}

As shown in Fig.~\ref{fig6}, our framework maintains stable object tracking on EndoVis2018, even with changes in target size, orientation, and position. MA-SAM2 relies on confidence scores and IoU values to maintain memory quality, which may filter out some less reliable masks under complex interactions (e.g., LND). Nevertheless, the model achieves consistent improvements across most categories, and we consider this a reasonable trade-off for overall robustness. Overall, our method demonstrates significant superiority in dynamic scenes compared to others, attributed to the memory-augmented strategy, thus ensuring excellent tracking precision.

\begin{table}
\centering
\caption{Ablation Study on EndoVis2017 and EndoVis2018 Datasets}
\begin{tabular}{c c c c c c c c c}
\hline
& \multicolumn{3}{c}{EndoVis2018} & & \multicolumn{3}{c}{EndoVis2017} \\
\hline
Method & Challenge IoU & IoU & mcIoU & & Challenge IoU & IoU & mcIoU \\
\hline
SAM2 & 60.04 & 60.04 & 58.40 & & 56.39 & 56.39 & 55.38 \\
+ CAM & 60.93 & 60.93 & 58.52 & & 57.86 & 57.86 & 56.33 \\
+ ORM & 63.98 & 63.98 & 60.31 & & 61.45 & 61.45 & 59.28 \\
+ CAM + ORM(Ours) & \textbf{64.40} & \textbf{64.40} & \textbf{62.13} & & \textbf{62.49} & \textbf{62.49} & \textbf{59.89} \\
\hline
\end{tabular}
\label{tab3}
\end{table}

\subsubsection{Ablation Study}
Table~\ref{tab3} demonstrates the critical contributions of the CAM and ORM modules to the overall performance of MA-SAM2. For both datasets, adding the CAM module to the baseline SAM2 model leads to an improvement in performance. Specifically, the Challenge IoU increases from 60.04\% to 60.93\% in EndoVis2018, showing that incorporating multi-pathway interaction enhances the model's feature representation. The further addition of our ORM module leads to a more substantial improvement in the Challenge IoU, reaching 64.40\% in the full model, confirming that this component helps further refine the segmentation results.

\section{Conclusion}
In this paper, we presented MA-SAM2, a training-free approach for instrument segmentation in surgical videos. It leverages a context-aware and occlusion-resilient memory model to address the limitations of SAM2 in surgical video segmentation. MA-SAM2 achieves robust and precise multi-target segmentation with a single prompt, effectively overcoming occlusions and disturbances caused by complex motions. Without requiring retraining, MA-SAM2 achieves a CIoU score of 64.40\% on the EndoVis2018 dataset and 62.49\% on the EndoVis2017 dataset, demonstrating its potential to enhance surgical quality.

\begin{credits}
\subsubsection{\discintname}
The authors have no competing interests.
\end{credits}

\bibliographystyle{splncs04}
\bibliography{Paper-2634}

\begin{thebibliography}{10}
\providecommand{\url}[1]{\texttt{#1}}
\providecommand{\urlprefix}{URL }
\providecommand{\doi}[1]{https://doi.org/#1}

\bibitem{Endovis2018}
Allan, M., Kondo, S., Bodenstedt, S., Leger, S., Kadkhodamohammadi, R., Luengo, I., Fuentes, F., Flouty, E., Mohammed, A., Pedersen, M., et~al.: 2018 robotic scene segmentation challenge. arXiv preprint arXiv:2001.11190  (2020)

\bibitem{Endovis2017}
Allan, M., Shvets, A., Kurmann, T., Zhang, Z., Duggal, R., Su, Y.H., Rieke, N., Laina, I., Kalavakonda, N., Bodenstedt, S., et~al.: 2017 robotic instrument segmentation challenge. arXiv preprint arXiv:1902.06426  (2019)

\bibitem{ayobi2023matis}
Ayobi, N., P{\'e}rez-Rond{\'o}n, A., Rodr{\'\i}guez, S., Arbel{\'a}ez, P.: Matis: Masked-attention transformers for surgical instrument segmentation. In: 2023 IEEE 20th International Symposium on Biomedical Imaging (ISBI). pp.~1--5. IEEE (2023)

\bibitem{ding2024sam2long}
Ding, S., Qian, R., Dong, X., Zhang, P., Zang, Y., Cao, Y., Guo, Y., Lin, D., Wang, J.: Sam2long: Enhancing sam 2 for long video segmentation with a training-free memory tree. arXiv preprint arXiv:2410.16268  (2024)

\bibitem{gonzalez2020isinet}
Gonz{\'a}lez, C., Bravo-S{\'a}nchez, L., Arbelaez, P.: Isinet: an instance-based approach for surgical instrument segmentation. In: International Conference on Medical Image Computing and Computer-Assisted Intervention. pp. 595--605. Springer (2020)

\bibitem{jian2020multitask}
Jian, Z., Yue, W., Wu, Q., Li, W., Wang, Z., Lam, V.: Multitask learning for video-based surgical skill assessment. In: 2020 Digital Image Computing: Techniques and Applications (DICTA). pp.~1--8. IEEE (2020)

\bibitem{kirillov2023segment}
Kirillov, A., Mintun, E., Ravi, N., Mao, H., Rolland, C., Gustafson, L., Xiao, T., Whitehead, S., Berg, A.C., Lo, W.Y., et~al.: Segment anything. In: Proceedings of the IEEE/CVF international conference on computer vision. pp. 4015--4026 (2023)

\bibitem{liu2024surgical}
Liu, H., Zhang, E., Wu, J., Hong, M., Jin, Y.: Surgical sam 2: Real-time segment anything in surgical video by efficient frame pruning. arXiv preprint arXiv:2408.07931  (2024)

\bibitem{ma2024segment}
Ma, J., He, Y., Li, F., Han, L., You, C., Wang, B.: Segment anything in medical images. Nature Communications  \textbf{15}(1), ~654 (2024)

\bibitem{ni2022surginet}
Ni, Z.L., Zhou, X.H., Wang, G.A., Yue, W.Q., Li, Z., Bian, G.B., Hou, Z.G.: Surginet: Pyramid attention aggregation and class-wise self-distillation for surgical instrument segmentation. Medical Image Analysis  \textbf{76},  102310 (2022)

\bibitem{peng2025sam}
Peng, Y., Lin, X., Ma, N., Du, J., Liu, C., Liu, C., Chen, Q.: Sam-lad: Segment anything model meets zero-shot logic anomaly detection. Knowledge-Based Systems p. 113176 (2025)

\bibitem{ravi2024sam}
Ravi, N., Gabeur, V., Hu, Y.T., Hu, R., Ryali, C., Ma, T., Khedr, H., R{\"a}dle, R., Rolland, C., Gustafson, L., et~al.: Sam2: Segment anything in images and videos. arXiv preprint arXiv:2408.00714  (2024)

\bibitem{shademan2016supervised}
Shademan, A., Decker, R.S., Opfermann, J.D., Leonard, S., Krieger, A., Kim, P.C.: Supervised autonomous robotic soft tissue surgery. Science translational medicine  \textbf{8}(337),  337ra64--337ra64 (2016)

\bibitem{shen2024performance}
Shen, Y., Ding, H., Shao, X., Unberath, M.: Performance and non-adversarial robustness of the segment anything model 2 in surgical video segmentation. arXiv preprint arXiv:2408.04098  (2024)

\bibitem{shvets2018automatic}
Shvets, A.A., Rakhlin, A., Kalinin, A.A., Iglovikov, V.I.: Automatic instrument segmentation in robot-assisted surgery using deep learning. In: 2018 17th IEEE international conference on machine learning and applications (ICMLA). pp. 624--628. IEEE (2018)

\bibitem{song2025attention}
Song, M., Zhai, C., Yang, L., Liu, Y., Bian, G.: An attention-guided multi-scale fusion network for surgical instrument segmentation. Biomedical Signal Processing and Control  \textbf{102},  107296 (2025)

\bibitem{videnovic2024distractor}
Videnovic, J., Lukezic, A., Kristan, M.: A distractor-aware memory for visual object tracking with sam2. arXiv preprint arXiv:2411.17576  (2024)

\bibitem{williams2024leaf}
Williams, D., Macfarlane, F., Britten, A.: Leaf only sam: A segment anything pipeline for zero-shot automated leaf segmentation. Smart Agricultural Technology  \textbf{8},  100515 (2024)

\bibitem{wu2023medical}
Wu, J., Ji, W., Liu, Y., Fu, H., Xu, M., Xu, Y., Jin, Y.: Medical sam adapter: Adapting segment anything model for medical image segmentation. arXiv preprint arXiv:2304.12620  (2023)

\bibitem{yan2024biomedical}
Yan, Z., Sun, W., Zhou, R., Yuan, Z., Zhang, K., Li, Y., Liu, T., Li, Q., Li, X., He, L., et~al.: Biomedical sam 2: Segment anything in biomedical images and videos. arXiv preprint arXiv:2408.03286  (2024)

\bibitem{yang2023track}
Yang, J., Gao, M., Li, Z., Gao, S., Wang, F., Zheng, F.: Track anything: Segment anything meets videos. arXiv preprint arXiv:2304.11968  (2023)

\bibitem{yang2022attention}
Yang, L., Gu, Y., Bian, G., Liu, Y.: An attention-guided network for surgical instrument segmentation from endoscopic images. Computers in Biology and Medicine  \textbf{151},  106216 (2022)

\bibitem{yang2023tma}
Yang, L., Wang, H., Gu, Y., Bian, G., Liu, Y., Yu, H.: Tma-net: A transformer-based multi-scale attention network for surgical instrument segmentation. IEEE Transactions on Medical Robotics and Bionics  \textbf{5}(2),  323--334 (2023)

\bibitem{yu2024sam}
Yu, J., Wang, A., Dong, W., Xu, M., Islam, M., Wang, J., Bai, L., Ren, H.: Sam 2 in robotic surgery: An empirical evaluation for robustness and generalization in surgical video segmentation. arXiv preprint arXiv:2408.04593  (2024)

\bibitem{yue2023cascade}
Yue, W., Liao, H., Xia, Y., Lam, V., Luo, J., Wang, Z.: Cascade multi-level transformer network for surgical workflow analysis. IEEE transactions on medical imaging  \textbf{42}(10),  2817--2831 (2023)

\bibitem{yue2024surgicalsam}
Yue, W., Zhang, J., Hu, K., Xia, Y., Luo, J., Wang, Z.: Surgicalsam: Efficient class promptable surgical instrument segmentation. In: Proceedings of the AAAI Conference on Artificial Intelligence. vol.~38, pp. 6890--6898 (2024)

\bibitem{zhang2020empowering}
Zhang, J., Tao, D.: Empowering things with intelligence: a survey of the progress, challenges, and opportunities in artificial intelligence of things. IEEE Internet of Things Journal  \textbf{8}(10),  7789--7817 (2020)

\bibitem{zhang2023personalize}
Zhang, R., Jiang, Z., Guo, Z., Yan, S., Pan, J., Ma, X., Dong, H., Gao, P., Li, H.: Personalize segment anything model with one shot. arXiv preprint arXiv:2305.03048  (2023)

\bibitem{zhu2024medical}
Zhu, J., Qi, Y., Wu, J.: Medical sam 2: Segment medical images as video via segment anything model 2. arXiv preprint arXiv:2408.00874  (2024)

\end{thebibliography}

\end{document}